\documentclass{article} 
\usepackage{iclr2026_conference,times}

\usepackage{amsmath,amsfonts,bm}

\def\eqref#1{equation~\ref{#1}}

\def\1{\bm{1}}

\DeclareMathAlphabet{\mathsfit}{\encodingdefault}{\sfdefault}{m}{sl}
\SetMathAlphabet{\mathsfit}{bold}{\encodingdefault}{\sfdefault}{bx}{n}

\usepackage{hyperref}
\usepackage{url}

\usepackage{booktabs}
\usepackage{graphicx}
\usepackage{multirow}
\usepackage{enumitem}

\title{EgoGapBench: Benchmarking Egocentric \\ Action Selection in Multi-Agent Scenes  }

\author{
Jihyeok Jung \\
KAIST AI \\
\texttt{ji9759@kaist.ac.kr} \\
\And
Jeewu Lee \\
Sogang University \\
\texttt{xlxaxb@sogang.ac.kr} \\
\And
Sanghyeop Kim \\
Sogang University \\
\texttt{fgd25@sogang.ac.kr} \\
\And
Chanhee Han \\
Ministry of Science and ICT \\
\texttt{enhance63@korea.kr} \\
\And
Seong Joon Oh\thanks{Corresponding author.} \\
KAIST AI \\
\texttt{coallaoh@kaist.ac.kr}
}

\iclrfinalcopy
\begin{document}

\maketitle
\lhead{Preprint}

\begin{abstract}

Existing egocentric benchmarks have primarily constructed the egocentric setting from first-person-view data, which makes it difficult to evaluate egocentric perspective itself in isolation. However, understanding first-person-view input and taking an egocentric perspective are separable abilities, especially when first-person body cues are absent or when other agents are present. To isolate egocentric perspective understanding, we introduce EgoGapBench, a diagnostic benchmark for measuring action selection in multi-agent egocentric scenes. We define the ability measured by this benchmark as Egocentric Action Selection (EAS): selecting an appropriate action from the agent's perspective in the presence of other agents. On EgoGapBench, humans answer reliably, whereas both open-source and proprietary MLLMs perform substantially worse and systematically select actions performed by other visible agents. Fine-tuning on existing egocentric data fails to close this gap and can even be detrimental. In contrast, fine-tuning on EgoGapBench training data improves accuracy but does not reach human performance. These results show that EAS is difficult to acquire from first-person-view data alone, and that MLLMs should be evaluated and trained not only for scene understanding but also for egocentric action selection. The code and benchmark are available at~\url{https://github.com/jhCOR/EgoGapBench}.
\end{abstract}

\section{Introduction}
\label{sec:intro}

\begin{figure}[t]
\centering
\includegraphics[width=\linewidth]{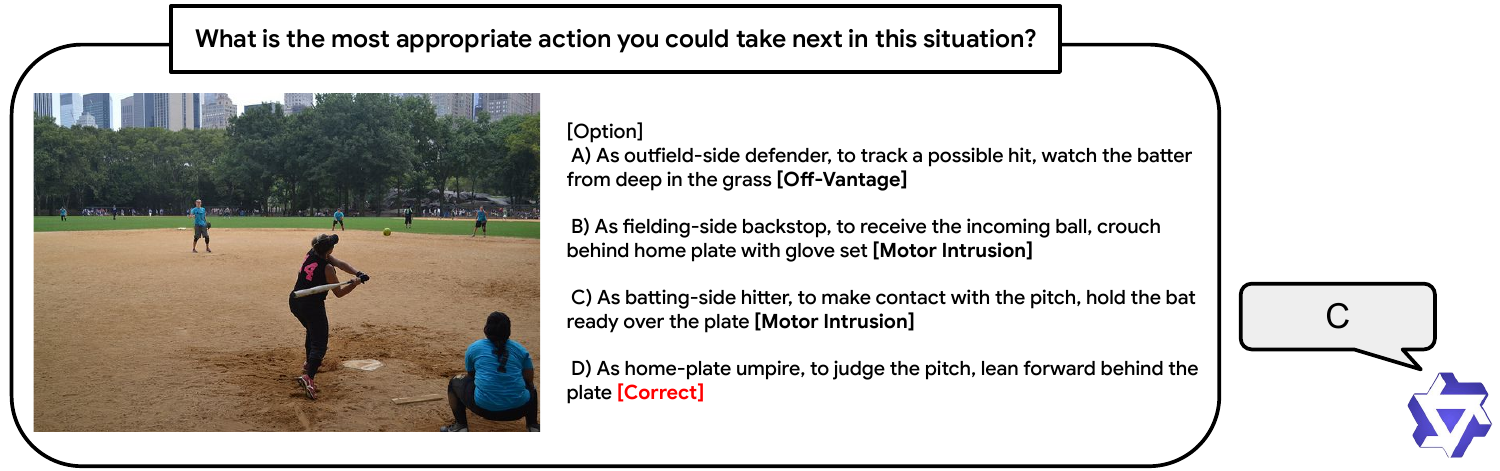}
\caption{An example item from EgoGapBench. The item asks the model to adopt a viewpoint behind home plate, corresponding to the umpire's role, in a scene that also contains other agents. Besides the correct answer, the options include the observed action of another agent (the batter preparing to hit, marked \emph{motor intrusion}) and a role that fits the activity but cannot be occupied from the given viewpoint (\emph{off-vantage}). Qwen3.5 selects a motor-intrusion option (C), choosing the action of the visible batter rather than an action appropriate from its assigned viewpoint.}
\label{fig:example}
\end{figure}

\begin{figure}[t]
\centering
\includegraphics[width=\linewidth]{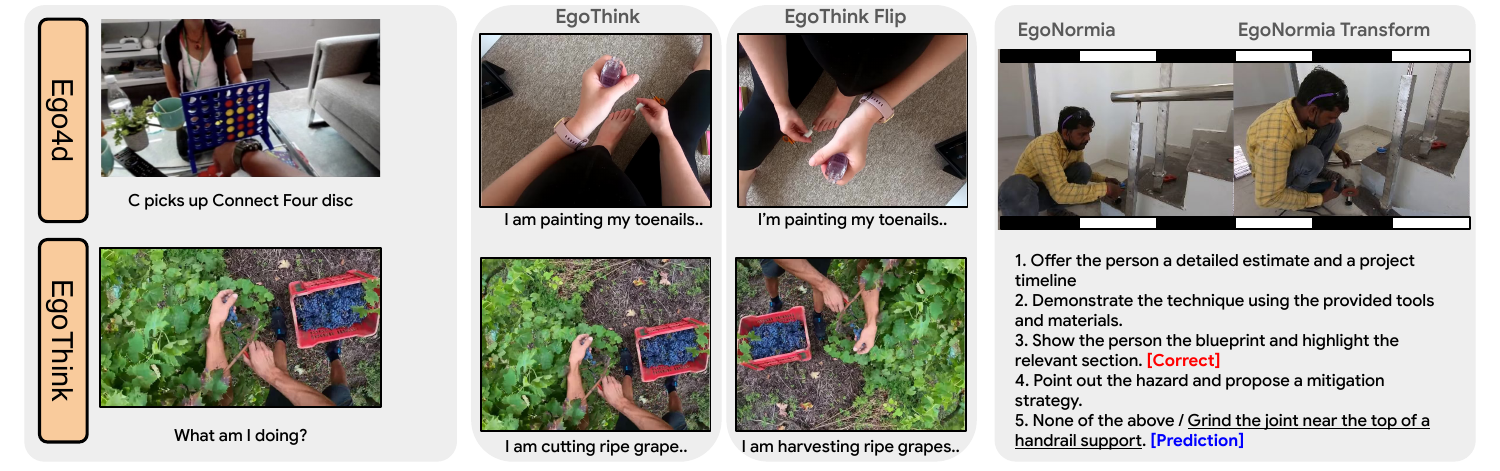}
\caption{ \textbf{Left:} many egocentric datasets and
benchmarks include visible camera-wearer body cues, allowing models to answer
from body evidence rather than perspective. \textbf{Middle:} vertically flipping an EgoThink image yields a viewpoint that could not arise from a first-person position, yet
models return nearly the same answer, indicating reliance on visible body cues.
 \textbf{Right:} on EgoNormia, replacing one option with an action observed in
the scene draws the model toward it (``Grind the joint near the top of a handrail support.'') over the action it
should take.}
\label{fig:teaser}
\end{figure}

Existing egocentric benchmarks~\citep{mangalam2023egoschema,rezaei2025egonormia,pan2026egointent, jia2022egotaskqa} have primarily constructed the egocentric setting using first-person-view data~\citep{grauman2022ego4d,damen2020epic}, making it difficult to isolate egocentric perspective itself~\citep{cheng2024egothink,jung2025right} (Figure~\ref{fig:teaser}, left). When we vertically flip images sampled from EgoThink~\citep{cheng2024egothink}, the model returns nearly the same answer to questions asking about the camera wearer's action, suggesting that it relies on observed first-person body cues (Figure~\ref{fig:teaser}, middle). Therefore, in existing egocentric benchmarks, it is difficult to verify whether a model's correct answer reflects a genuine understanding of egocentric perspective or merely the reading of visible body cues. However, as shown in Figure~\ref{fig:example}, current MLLMs, unlike humans, struggle to choose appropriate actions when first-person body cues are absent~\citep{zhang2026mindpower}.

In this study, we distinguish the ability to understand first-person-view input~\citep{grauman2022ego4d} from the ability to understand an egocentric perspective. The former concerns the ability to interpret visual input captured from the field of view of an agent situated in the scene. In contrast, the latter concerns the ability to understand a scene from the observer's standpoint, and is therefore not limited to first-person-view input but can also apply to more general input formats~\citep{jung2025right}. The two abilities usually operate together, but they can become separable when the agent's body is not visible or when other agents are present in the scene. This separation becomes especially important in action selection: when visible first-person cues are unavailable or another agent's action is visually salient, a model may select an incorrect action. Indeed, when another agent's observed action is added as a candidate option to items sampled from EgoNormia's~\citep{rezaei2025egonormia} norm-aligned action-selection task, even Gemini-3-Flash~\citep{deepmind2025gemini3flash} is drawn toward that action and its accuracy drops by nearly half (Figure~\ref{fig:teaser}, right; Appendix~\ref{app:pilot}).

To isolate egocentric perspective understanding, we propose EgoGapBench, a benchmark for measuring action selection in multi-agent egocentric scenes without visible first-person-view cues. We refer to the underlying ability as Egocentric Action Selection (EAS), defined as selecting an appropriate action from the agent's perspective in the presence of other agents. EgoGapBench consists of 1{,}000 action-selection items built from multi-agent scenes collected from a general image pool without first-person body cues. The model cannot rely on the camera wearer's visible body or first-person distortion, and other agents' actions serve as plausible but perspective-inappropriate attractors. Real scenes often allow more than one appropriate action, and an agent may sometimes follow another person's action without being clearly wrong. To avoid this ambiguity, we construct incorrect options using singleton actions: actions that cannot be duplicated by multiple agents in the same situation, such as the batter's action in Figure~\ref{fig:example}. This design makes it possible to measure whether the model selects an appropriate action, because copying another agent's action leads to a clear error.

Evaluated on EgoGapBench, both open-source and proprietary models perform far below humans, and they systematically select actions performed by other agents more often, especially when the correct answer is an active action rather than a passive action such as observing. Fine-tuning on existing egocentric data does not help and can even hurt performance, whereas fine-tuning on a split of EgoGapBench improves accuracy but does not close the gap with human performance. Overall, EAS is difficult to acquire from first-person-view data alone and calls for structural improvements that can enhance egocentric action selection.

\paragraph{Contributions.}
We make three contributions:
\begin{itemize}
\setlength{\itemsep}{2pt}
\item We formalize Egocentric Action Selection (EAS) by separating egocentric perspective understanding from first-person-view understanding.
\item We introduce EgoGapBench, a 1{,}000-item diagnostic benchmark for evaluating EAS in body-free multi-agent scenes, where singleton-action distractors make copying another agent's observed action a clear error.
\item Through evaluations of open-source and proprietary MLLMs, we show a large human--model gap and a systematic tendency to copy visible others' actions. We further find that existing egocentric data does not reliably induce EAS, while EAS-specific tuning improves performance but leaves structured errors.
\end{itemize}

\section{Related Work}
\label{sec:related}

\subsection{Egocentric Scene Understanding}
\label{subsec:rel-understanding}

Existing egocentric benchmarks have greatly advanced first-person view understanding. Ego4D~\citep{grauman2022ego4d} and EPIC-KITCHENS~\citep{damen2020epic} provide large-scale first-person view data for recognizing activities, objects, and interactions, and for forecasting future actions. Based on these data, EgoThink~\citep{cheng2024egothink} evaluates perception and reasoning in first-person views, while EgoSchema~\citep{mangalam2023egoschema} evaluates long-video understanding. These benchmarks require models to understand what the camera wearer is doing, what the wearer is seeing, and what the wearer is likely to do next. EgoGapBench complements this line of work by shifting the focus from first-person view understanding to egocentric action selection. Specifically, EgoGapBench uses scenes without common first-person visual cues and evaluates whether a model can take the given viewpoint as an acting perspective in multi-agent scenes.

\subsection{Action Selection in MLLMs}
\label{subsec:rel-action}

Recent MLLM benchmarks increasingly evaluate whether models can select actions from visual input, but existing action-selection tasks do not make another agent's action a competing candidate for the model's own action. EgoPlan-Bench~\citep{chen2026egoplan} evaluates action planning from first-person video, and EgoNormia~\citep{rezaei2025egonormia} evaluates norm-appropriate action selection. However, their candidate actions are generally framed as actions available to the camera wearer. Even when other people appear in the scene, they usually serve as social or physical context for the wearer's action, rather than as agents whose own actions must be rejected because they do not belong to the viewpoint agent. EgoGapBench targets this diagnostic case by including another agent's observed action as a structurally incorrect option, thereby testing whether a model can distinguish actions available from its own viewpoint from actions performed by others.

\subsection{Self-Other Distinction in Vision-Language Models}
\label{subsec:rel-sod}

Recent work suggests that egocentric input alone does not guarantee a stable distinction between the camera wearer and other people. IndEgo~\citep{chavan2026indego} and MyEgo~\citep{xiao2026ego} study whether models can ground the camera wearer, the wearer's body, objects, or personal context, and report failures of egocentric grounding in which models fail to separate the wearer from others. One line of work addresses this problem by explicitly providing such a distinction. MindPower~\citep{zhang2026mindpower} improves embodied reasoning by prompting the model to act as a helpful robot, thereby textually separating the model's agent identity from the humans in the scene. EgoGapBench asks whether a model can take the given visual viewpoint as its own action perspective and choose an action available from that viewpoint when self-other separation is not directly specified.


\section{EgoGapBench}
\label{sec:benchmark}

\subsection{Task Definition}
\label{subsec}

Given an image of a multi-agent scene, the EAS task presents a question and a set of candidate actions. The model must treat the given viewpoint as its own and select one action that is immediately available from that viewpoint. An answer is correct only if the selected action is appropriate for the agent occupying the viewpoint, rather than for another visible agent in the scene.

The candidate options include both active interventions and passive responses, such as watching or waiting. Incorrect options are constructed to test two distinct failures. The first is choosing a singleton action that another visible agent is already performing. The second is choosing an \emph{off-vantage} action, whose action may fit the activity but whose position cannot be occupied from the given viewpoint. This tests whether the model can localize itself correctly in the scene, rather than merely choose a plausible scene-level action.

Each item is evaluated in two formats. The three-option format (n3) contains one correct answer and two incorrect options based on actions performed by other visible agents. The four-option format (n4) adds one off-vantage incorrect option.

\subsection{Design Principles}
\label{subsec:design}

\paragraph{Motivation.}
To evaluate egocentric action selection in isolation, EgoGapBench must separate two factors that are often entangled in existing first-person benchmarks: recognizing first-person visual cues and selecting an action from the assigned viewpoint. First-person footage can expose body-based cues such as hands, feet, camera height, or capture distortion, allowing a model to identify the camera wearer without reasoning about the viewpoint as an action standpoint. At the same time, scenes without other agents provide little evidence for whether the model distinguishes its assigned viewpoint from another agent's role. EgoGapBench therefore uses multi-agent scenes without visible first-person-specific cues, so that other agents serve as contrastive evidence for self-other separation while the viewpoint itself must be used to determine the available action.

We operationalize this idea through two design principles.

\paragraph{Principle 1: Singleton constraint.}
We construct distractors from singleton actions: actions that cannot be duplicated by multiple agents in the same situation. Under this constraint, an option derived from another agent's observed action is structurally incorrect rather than merely less plausible. For example, in a baseball scene, if another visible agent is already the batter, the viewpoint agent cannot also take the batter's action. This makes copying another agent's action a diagnostic error for self-other confusion.

\paragraph{Principle 2: No first-person-specific cues.}
We construct the benchmark from ordinary images that do not contain visible first-person-specific cues, such as the camera wearer's hands or feet, low camera height, or lens distortion. This prevents the task from being solved by detecting first-person visual artifacts. Instead, the model must relate the assigned viewpoint to the action possibilities in the scene.

\subsection{Dataset Construction}
\label{subsec:construction}

To implement the two principles above, we select singleton scenes from COCO
train 2017~\citep{lin2014microsoft}, generate options for each scene, verify them
with human annotators, and assemble the final items. Most stages are automated,
so the same procedure can scale to a larger collection. Per-stage counts and
models are given in Appendix~\ref{app:datapipe}.

\paragraph{Image selection.}
Ordinary image datasets do not annotate whether the roles or actions of visible agents are mutually exclusive in a given scene. We therefore use an LLM to identify candidate singleton structures: objects, roles, people, or actions that are difficult to duplicate in the same situation. For example, a bride and groom in a wedding scene, or the ball carrier in a soccer game, can define such singleton roles.

Starting from COCO images that contain at least two people, we extract candidate images by checking whether two  agents independently satisfy the singleton constraint. Two human annotators then review each image to determine whether the agents' roles form a valid singleton structure and whether the associated action descriptions match the scene.

\paragraph{Option generation.}
For each collected image, an LLM generates a candidate correct action that is available from the assigned viewpoint. It also generates an \emph{off-vantage} action: an action that may fit the overall activity, but cannot be performed immediately from the assigned viewpoint without changing position. For example, an action available to a player on the field is off-vantage when the assigned viewpoint is located in the spectator seats. Correct actions are divided into two types: \emph{direct}, where the appropriate response is active, and \emph{indirect}, where the appropriate response is passive, such as watching or listening. Incorrect options based on other agents' observed actions are not newly generated at this stage; instead, we reuse the observed actions identified during image selection and rewrite them into the same format as the other options.

\paragraph{Annotation.}
Two annotators independently label each generated correct-action candidate as a valid correct answer, an off-vantage option, or discard. For candidates generated as off-vantage options, annotators label each candidate as either valid off-vantage or discard. We retain only candidates for which both annotators assign the same label. In particular, off-vantage candidates that are overly awkward in the scene or that simply copy a visible agent's action are discarded. For correct-action candidates, the two-way agreement distinguishing usable options (valid correct answers) from unusable ones (off-vantage and discard combined) is $\kappa=0.814$, and the three-way agreement over valid correct answer / off-vantage option / remove is $\kappa=0.61$~(Appendix~\ref{app:iaa}).

\paragraph{Item assembly.}
Each verified correct answer becomes one item. To prevent items from the same image from appearing in both training and evaluation, we split the data at the image level for the fine-tuning analysis (Section~\ref{subsec:tuning}). Each item is paired
with incorrect options built from the observed actions of other agents in the
scene, and n4 items additionally include one off-vantage option. The final benchmark consists of 1{,}000 items built from 440
images, with 650 direct and 350 indirect items, and 618 n3 and 382 n4 items.
Detailed statistics for candidate labels are given in Appendix~\ref{app:datapipe}.

\paragraph{Out-of-domain set.}
To test whether the same gap persists outside the COCO domain, we apply the same pipeline to 44 manually collected Wikimedia images and build an out-of-domain set of 107 items (Appendix~\ref{app:wiki}).

\section{Experiments}
\label{sec:experiments}

\subsection{Setup}
\label{subsec:setup}

\paragraph{Evaluation.}
EgoGapBench is evaluated in three-option (n3) and four-option (n4) multiple-choice formats. We compute accuracy by string-matching the model's predicted option against the ground-truth answer. In addition to overall accuracy, we report accuracy separately for direct and indirect items, according to whether the correct action requires an active intervention or a passive response. For n4 items, we further analyze error types by reporting how often models select the observed-action option, which we call a motor-intrusion error, and how often they select the off-vantage option.

\paragraph{Models.}
We evaluate open-source and proprietary MLLMs zero-shot. The open-source models
are gemma-4-E4B-it~\citep{google2026gemma4}, Qwen2.5-VL-7B-Instruct~\citep{qwen2.5-VL}, Qwen3.5-9B~\citep{qwen3.5}, InternVL2.5-8B~\citep{chen2024expanding}, and 
InternVL3.5-8B~\citep{wang2025internvl3_5}. The
proprietary models are gpt-5.4-2026-03-05~\citep{openai2026gpt54}, gpt-4o-2024-11-20~\citep{hurst2024gpt}, and gemini-3-flash-preview~\citep{deepmind2025gemini3flash}. Fine-tuning
experiments are conducted only on the five open-source models. The human baseline
is measured by having one annotator, who did not participate in benchmark
construction, answer 200 randomly sampled items.

\paragraph{Fine-tuning.}
We compare two fine-tuning regimes: egocentric tuning on the existing
egocentric instruction dataset EgoIT-99K~\citep{yang2025egolife}, and EAS tuning on the EgoGapBench training split with short LLM-generated rationales. Both use the same LoRA
recipe via MS-Swift~\citep{zhao2024swiftascalablelightweightinfrastructure}; data scale and hyperparameters are in
Appendix~\ref{app:training}.

\subsection{Models Fail at Egocentric Action Selection}
\label{subsec:gap}

\paragraph{Result on EgoGapBench.}
As shown in Table~\ref{tab:zeroshot}, current MLLMs perform far below humans on EgoGapBench. Humans reach $94.5\%$ accuracy, whereas the strongest model, GPT-5.4 (medium), reaches only $66.1\%$ overall, and most open-source models remain near chance. The InternVL models fall below chance. These results indicate that current MLLMs cannot reliably choose the action available to the agent occupying the given viewpoint in multi-agent scenes.

The gap is especially pronounced for items whose correct answer is a \emph{direct} action. Across all models, accuracy on direct-action items is lower than accuracy on indirect-action items. Even GPT-5.4 (medium) reaches only $50.9\%$ on direct items, compared with $94.3\%$ on indirect items. This suggests that recognizing the overall situation and selecting what one should do within that situation are separable abilities. At the same time, the failure is not limited to active action selection: for example, GPT-4o reaches only $52.3\%$ even on indirect items, suggesting that some models remain strongly influenced by other agents in the scene.

The n3 and n4 results further suggest that the failure is not simply caused by the number of options or by the off-vantage distractor. Although n4 adds one off-vantage option to the n3 format, many models show only small accuracy differences between the two formats; for example, Gemma4 scores $53.4\%$ on n3 and $49.0\%$ on n4, and InternVL2.5 performs higher on n4 than on n3. The error breakdown in the right block of Table~\ref{tab:zeroshot} shows that, on n4 items, errors instead concentrate on the observed-action option. Thus, models are not failing at random, but are systematically attracted to actions already performed by other agents in the scene. This attraction weakens as model capability increases, but it does not disappear.

Taken together, these results suggest that the failure on EgoGapBench is not merely a failure of scene understanding. Models often recognize actions occurring in the scene, but fail to judge which action is available from the assigned egocentric standpoint, instead tending to copy or follow another agent's action.

\begin{table}[t]
\centering
\caption{Zero-shot results on EgoGapBench (\%), sorted by overall accuracy.
\textbf{Left:} accuracy---even the strongest proprietary models score high on
\emph{indirect} items but collapse on \emph{direct} items. \textbf{Right:} n4
errors split by type. Each value is reported as a share of \emph{all} n4 items,
so that n4 accuracy and the two error columns partition the n4 set (values are
rounded to one decimal place). Except for three rows, motor-intrusion errors
exceed the random-choice baseline. Results are robust to prompt formulation
(Appendix~\ref{app:egoplan}).}

\label{tab:zeroshot}
\small
\setlength{\tabcolsep}{5pt}
\begin{tabular}{l ccc cc c cc}
\toprule
& \multicolumn{5}{c}{Accuracy (\%)} & & \multicolumn{2}{c}{n4 errors (\% of all n4)} \\
\cmidrule(lr){2-6} \cmidrule(lr){8-9}
Model & Overall & n3 & n4 & Direct & Indirect & & Motor intr. & Off-vantage \\
\midrule
Random            & 30.2 & 33.3 & 25.0 & 30.1 & 30.2 & & 50.0 & 25.0 \\
\midrule
gpt-5.4 (medium)  & 66.1 & 66.3 & 65.7 & 50.9 & 94.3 & & 21.5 & 12.8 \\
gpt-5.4 (low)     & 58.3 & 58.9 & 57.3 & 40.9 & 90.6 & & 32.7 & 10.0 \\
gemma4            & 51.7 & 53.4 & 49.0 & 34.0 & 84.6 & & 35.3 & 15.7 \\
gemini-3-flash (low)   & 39.2 & 42.7 & 33.5 & 27.8 & 60.3 & & 58.9 & \phantom{0}7.6 \\
qwen35            & 29.0 & 31.9 & 24.3 & 20.0 & 45.7 & & 69.4 & \phantom{0}6.3 \\
qwen25vl          & 28.6 & 29.6 & 27.0 & 21.8 & 41.1 & & 63.4 & \phantom{0}9.7 \\
gpt-4o            & 26.7 & 31.1 & 19.6 & 12.9 & 52.3 & & 73.0 & \phantom{0}7.3 \\
internvl35        & 24.1 & 25.1 & 22.5 & 20.5 & 30.9 & & 74.6 & \phantom{0}2.9 \\
internvl25        & 16.1 & 15.0 & 17.8 & 17.2 & 14.0 & & 77.8 & \phantom{0}4.5 \\
\midrule
Human$^\dagger$   & 94.5 & 98.4 & 88.2 & 93.2 & 97.1 & & \phantom{0}3.9 & \phantom{0}7.9 \\
\bottomrule
\end{tabular}
\\[2pt]
{\footnotesize $^\dagger$One annotator, who did not take part in data construction, evaluated all 200 items.}
\end{table}

\subsection{Effect of Egocentric Data Tuning}
\label{subsec:dissociation}

As shown in Table~\ref{tab:dissociation}, existing egocentric tuning lowers
EgoGapBench accuracy in all five models (mean $-33.9\%$). In contrast, its effect
on EgoThink, an external benchmark for first-person scene understanding, is
negligible on average. The effect is largest for Gemma4, which starts at
$51.7\%$ on EgoGapBench and drops to $18.4\%$ after egocentric tuning, below
chance.

This result shows that existing egocentric tuning does not reliably build the ability to choose an appropriate action from the given viewpoint. The fact that the same tuning leaves EgoThink scene-understanding performance almost unchanged while substantially lowering EgoGapBench performance suggests that first-person scene understanding and action selection from the given viewpoint can behave differently. Existing first-person data often contains body cues such as hands, feet, and manipulated objects, which may allow a model to infer the answer from such cues without using the given viewpoint as the standpoint for its own action. The fact that models tuned on such data fail more strongly on EgoGapBench, where body cues are absent, suggests that existing egocentric tuning may not provide a direct learning signal for EAS.

\begin{table}[t]
\centering
\caption{\textbf{Egocentric tuning on EgoIT-99K} dissociates Egocentric Action
Selection (EgoGapBench) from first-person scene understanding (EgoThink): the
same training leaves scene parsing essentially unchanged (mean $-0.4\%$) while
collapsing EgoGapBench across all five models (mean $-33.9\%$). EgoThink is the
mean of $\{0,0.5,1\}$ judge grades (GPT-4o); EgoGapBench is multiple-choice
accuracy (\%, chance $\approx$30\%). As the two scales are not directly
comparable in raw points, we contrast them via self-normalized relative change
(rel.). Columns: ZS (zero-shot) and EgoIT (after egocentric tuning on EgoIT-99K).}
\label{tab:dissociation}
\small
\setlength{\tabcolsep}{6pt}
\begin{tabular}{l ccc c ccc}
\toprule
& \multicolumn{3}{c}{\textbf{EgoThink} (0--1)} & & \multicolumn{3}{c}{\textbf{EgoGapBench} (acc.\ \%)} \\
\cmidrule(lr){2-4} \cmidrule(lr){6-8}
Model & ZS & EgoIT & rel. & & ZS & EgoIT & rel. \\
\midrule
gemma4     & 0.533 & 0.543 & $+1.9\%$ & & 51.7 & 18.4 & $-64.4\%$ \\
qwen35     & 0.653 & 0.676 & $+3.5\%$ & & 29.0 & 17.4 & $-40.0\%$ \\
qwen25vl   & 0.631 & 0.596 & $-5.6\%$ & & 28.6 & 23.8 & $-16.8\%$ \\
internvl35 & 0.626 & 0.608 & $-2.9\%$ & & 24.1 & 21.7 & $-10.0\%$ \\
internvl25 & 0.590 & 0.595 & $+0.8\%$ & & 16.1 & \phantom{0}9.9 & $-38.5\%$ \\
\midrule
\textbf{Mean} & 0.606 & 0.603 & $\mathbf{-0.4\%}$ & & 29.9 & 18.2 & $\mathbf{-33.9\%}$ \\
\bottomrule
\end{tabular}
\end{table}

\subsection{Controlled Analyses}
\label{subsec:tuning}

\paragraph{Prompt variation.}
We first rephrase the prompt into an EgoPlan~\citep{chen2026egoplan} style prompt that asks for the camera wearer's appropriate action. This prompt generally preserves the trend across models: Spearman $\rho$ is $0.60$, three of the five models keep their rank, and only the two near-chance models swap places. This suggests that the performance gap observed on EgoGapBench does not depend strongly on a particular prompt phrasing (Appendix~\ref{app:egoplan}).

We further test a prompt that explicitly instructs the model not to choose an
action another person is performing. This instruction improves
accuracy while largely preserving the model ranking, confirming that models are
affected by observed actions of other agents. However, performance still remains
far below the human baseline, suggesting that directly stating the exclusion rule
does not solve viewpoint-grounded action selection (Appendix~\ref{app:observeself}).

\paragraph{Out-of-domain Data Evaluation.}
To test whether the results depend only on the COCO image domain, we also
evaluate on an out-of-domain set built by applying the same construction pipeline
to manually collected Wikimedia images. On this set, open-source models again
perform poorly in the zero-shot setting, and direct items remain harder than
indirect items. This suggests that the failure measured by EgoGapBench is not
limited to a particular image source or to the scene distribution of COCO
(Appendix~\ref{app:wiki}).

\paragraph{Effect of EAS-specific tuning.}
As shown in Table~\ref{tab:rationale}, EAS data tuning substantially improves accuracy but does not close the gap to human performance. For example, Qwen3.5 improves from $29.0\%$ to $75.2\%$ overall, and from $20.0\%$ to $63.4\%$ on direct items. The model ranking is also largely preserved relative to the zero-shot setting (Spearman $\rho = 0.90$ over the five open-source models), suggesting that EgoGapBench scores are not merely an incidental format effect but reflect relatively stable differences across models.

However, the remaining errors are still structured. In the right block of Table~\ref{tab:rationale}, motor-intrusion errors decrease for all five models, while off-vantage errors increase as a share of all n4 items. For Qwen3.5, motor-intrusion errors drop from $69.4\%$ to $14.9\%$, whereas off-vantage errors rise from $6.3\%$ to $23.0\%$. This suggests that EAS data tuning reduces the tendency to copy another visible agent's action, but does not fully solve viewpoint-grounded action selection; instead, some errors shift from motor intrusion to off-vantage mislocalization. The InternVL2.5 models also show limited gains after tuning, reaching only $41.3\%$ and $38.6\%$ overall.

Taken together, the gap persists across prompt variations and out-of-domain data,
so it is not an artifact of prompt format or the COCO domain. Attraction to
another agent's action emerges as a central failure mode, yet even after EAS
tuning a large gap to humans and structured errors remain, indicating that EAS is
not fully solved by prompting or same-format training.

\begin{table}[t]
\centering
\caption{\textbf{EAS tuning} (fine-tuning on EAS data to match the
evaluation format) partially recovers EgoGapBench accuracy over zero-shot
(Table~\ref{tab:zeroshot}). The right block reports the direction of n4 errors as
a share of \emph{all} n4 items, before (ZS, derived from Table~\ref{tab:zeroshot})
and after EAS tuning. Although accuracy improves, in all five models
motor-intrusion errors fall while off-vantage errors rise. Accuracy (\%), open-source models only.}
\label{tab:rationale}
\small
\setlength{\tabcolsep}{5pt}
\begin{tabular}{l ccc cc cc cc}
\toprule
& \multicolumn{5}{c}{Accuracy (\%)} & \multicolumn{4}{c}{n4 error direction (\% of n4)} \\
\cmidrule(lr){2-6} \cmidrule(lr){7-10}
& & & & & & \multicolumn{2}{c}{Motor intr.\ } & \multicolumn{2}{c}{Off-vantage} \\
\cmidrule(lr){7-8} \cmidrule(lr){9-10}
Model & Overall & n3 & n4 & Direct & Indirect & ZS & EAS & ZS & EAS \\
\midrule
qwen35     & 75.2 & 83.3 & 62.0 & 63.4 & 97.1 & 69.4 & 14.9 & \phantom{0}6.3 & 23.0 \\
gemma4     & 56.9 & 61.2 & 50.0 & 38.6 & 90.9 & 35.3 & 31.7 & 15.7 & 18.3 \\
qwen25vl   & 54.1 & 58.1 & 47.6 & 40.8 & 78.9 & 63.4 & 34.3 & \phantom{0}9.7 & 18.1 \\
internvl35 & 41.3 & 43.9 & 37.2 & 29.7 & 62.9 & 74.6 & 53.8 & \phantom{0}2.9 & \phantom{0}8.9 \\
internvl25 & 38.6 & 39.3 & 37.4 & 29.2 & 56.0 & 77.8 & 51.3 & \phantom{0}4.5 & 11.3 \\
\bottomrule
\end{tabular}
\end{table}
\section{Conclusion}
\label{sec:conclusion}

We distinguished first-person-view understanding from the ability to take an egocentric perspective, and studied one concrete form of it, Egocentric Action Selection (EAS). To this end, we
presented EgoGapBench, a diagnostic benchmark for measuring whether models can
perform EAS without relying on visible first-person cues or another agent's
observed action. EgoGapBench evaluates whether a model can ground action
selection in the given viewpoint.

Evaluating open-source and proprietary MLLMs, we find that humans answer
reliably, whereas current models perform far worse. Models do not fail at random:
they are systematically drawn to the observed action of another agent, especially
when the correct answer requires an active action from the given viewpoint.
Existing egocentric tuning does not close this gap and can even degrade
EgoGapBench performance while largely preserving first-person scene understanding.

These results show that understanding first-person visual content does not by
itself guarantee viewpoint-grounded action selection. Improving EAS may require
models to go beyond recognizing camera-wearer cues or visible actions, and to
learn to choose actions from the given viewpoint. We hope EgoGapBench serves as a
diagnostic tool for measuring progress toward this goal.

\subsubsection*{Acknowledgments}
This work was supported by Institute for Information \& communications Technology Planning \& Evaluation(IITP)grant funded by the Korea government(MSIT) (RS-2019-II190075, Artificial Intelligence Graduate School Program(KAIST))

\bibliography{iclr2026_conference}
\bibliographystyle{iclr2026_conference}

\appendix
\section{Limitations}
\label{app:limitations}

A natural follow-up is to stratify items by the relation between agents, such as
independent, collaborative, or hindering. Such a taxonomy, however, does not fit
the measurement frame of EAS. Independent situations cannot be measured at all:
if an action is unrelated to the other agent, any action can be correct, so a
single-answer item does not exist. Collaboration and hindering do not separate
cleanly either. A sports scene is collaborative at the level of making the game
possible, yet hindering at the level of denying the opponent a win, and the same
item belongs to both depending on the level one reads it at. We therefore do not
adopt relation type as a first-class axis: doing so would pay the cost of
ill-defined categories that outweighs, against this benchmark's aim of diagnostic
clarity, what it would add. Our direct/indirect distinction separates action
selection from mere attribution without relying on this relational structure.
Finally, we measure self-other confusion at the motor level, as the copying of an
observed action; confusion at the level of goals, where a model adopts another
agent's objective as its own, is left as future work that could complement this
diagnostic.
\section{Motivating Pilot: Sensitivity to the Observed Action}
\label{app:pilot}

The observation in Section~\ref{sec:intro} comes from a small pilot on a
third-party benchmark, EgoNormia, which we report here in full. It suggests that models are drawn
to an action observed in the scene, the tendency our benchmark is designed to
measure directly.

\paragraph{Construction.}
We start from EgoNormia benchmark, which includes per-frame descriptions of what happens in the scene. Using GPT-5.4, we automatically select items in which (i) an agent other than the viewer is present and (ii) the final-frame description provided by the benchmark refers to an action that is not the viewer's own, yielding N = 543 items. For each item, we compare three cases: the original fifth option ("none of the above"), the case where it is replaced with an action drawn from a random other video, and the case where it is replaced with an action actually observed in the same scene, taken from the per-frame description.

\paragraph{Results.}
Table~\ref{tab:pilot} reports accuracy and the rate at which the fifth option is
chosen (Pick5). When the added option is a random action, models rarely choose
it (Pick5 10.1\% for gpt-4o-2024-11-20 and 9.8\% for gemini-3-flash-preview). When it is the action observed
in the scene, Pick5 rises sharply, to 35.0\% for GPT-4o and 51.6\% for Gemini,
and accuracy falls (GPT-4o $56.7 \rightarrow 42.4$, Gemini $66.9 \rightarrow
32.4$). Because the shuffled control does not produce this effect, the models are
drawn to the observed action in particular, consistent with the motor intrusion
we study with EgoGapBench.

\begin{table}[t]
\centering
\caption{Motivating pilot on EgoNormia ($N=543$). Adding a fifth option that is
an action observed in the scene (video\_action) sharply raises how often models
choose it (Pick5) and lowers accuracy, whereas a random action (shuffled) does
not. Pick5 is the percentage of items on which the added option is selected.}
\label{tab:pilot}
\small
\setlength{\tabcolsep}{8pt}
\begin{tabular}{l cc cc}
\toprule
& \multicolumn{2}{c}{GPT-4o} & \multicolumn{2}{c}{Gemini} \\
\cmidrule(lr){2-3} \cmidrule(lr){4-5}
Condition (5th option) & Acc(\%) & Pick5(\%) & Acc(\%) & Pick5(\%) \\
\midrule
original (none-of-above)      & 56.7 & 11.0 & 66.9 & \phantom{0}2.6 \\
shuffled (random video action) & 57.5 & 10.1 & 62.6 & \phantom{0}9.8 \\
video\_action (this video's action) & 42.4 & 35.0 & 32.4 & 51.6 \\
\bottomrule
\end{tabular}
\end{table}
\section{Prompts}
\label{app:prompt}

We use two prompt styles. The \emph{EgoGapBench} prompt addresses the model in the
second person (``you''), placing it as the agent present in the scene. The
\emph{egoplan}~\citep{chen2026egoplan} style prompt is a perturbation used for robustness checks
(Appendix~\ref{app:egoplan}): it keeps the same premises but rephrases the
question in the first person (``I'') and adds standard multiple-choice phrasing,
referring to the viewpoint as the camera wearer's. Model rankings are generally
preserved across the two styles.

\paragraph{EgoGapBench prompt.}
\begin{quote}\small\ttfamily
Imagine that the depicted scene is happening in front of you right now.
You are physically present at this location and directly seeing the situation
with your own eyes.\\[2pt]
Your task is to choose the most appropriate action you could take next in this
situation.\\[2pt]
Note: An appropriate action must be possible right now, without needing to walk,
run, relocate, spend time moving, or wait for a turn.\\[2pt]
Options: \{options\_block\}\\[2pt]
Respond with compact JSON and nothing else:
\{ "answer": \{answer\_letters\_spec\}, "rationale": "<=25 words" \}
\end{quote}

\paragraph{EgoPlan Style prompt.}
\begin{quote}\small\ttfamily
Imagine that the depicted scene is happening in front of you right now.
You are physically present at this location and directly seeing the situation
with your own eyes.\\[2pt]
Select the best answer to the following multiple-choice question based on the
image.\\[2pt]
Considering my current observation in the image, what action should I take
next?\\[2pt]
Note: An appropriate action must be possible right now, without needing to walk,
run, relocate, spend time moving, or wait for a turn.\\[2pt]
Behavior options: \{options\_block\}\\[2pt]
Respond with compact JSON and nothing else:
\{ "answer": \{answer\_letters\_spec\}, "rationale": "<=25 words" \}
\end{quote}
\section{Data Construction Pipeline}
\label{app:datapipe}
This appendix details the image-selection pipeline summarized in
Section~\ref{subsec:construction}. The goal at every stage is to retain singleton
images, in which two agents in the scene each have a distinct, single action,
while keeping the procedure automatic enough to scale.

\paragraph{Stage 1: annotation-based filter.}
From the 118{,}287 images of COCO train2017, we use the instance and keypoint
annotations alone, with no model, to keep images that contain two or more clearly
depicted non-crowd people. From the keypoint annotations, we then keep only
images in which at least two of these people have enough visible keypoints
(at least eight) for their action to be identifiable rather than being small
background figures. This stage leaves 15{,}446 images.

\paragraph{Stage 2: VLM singleton filter.}
We then filter for the singleton property with two vision-language models in
sequence, extracting the singleton and the actions of the agents that hold the
singleton property. We first use Qwen3.5-35B-A3B~\citep{qwen3.5} and then the
gpt-5.4-mini-2026-03-17 API~\citep{openai2026gpt54}, each prompted to judge whether the scene contains
agents in distinct, non-duplicable roles. The two stages leave 7{,}925 and
then 2{,}518 images, respectively. For images manually collected from Wikimedia,
however, we use only gpt-5.4-2026-03-05 for filtering.

\paragraph{Stage 3: category balancing.}
Because COCO can be dominated by a few activities (for example, baseball), we
classify each surviving image into a short activity category using
Qwen3.5-35B-A3B and cap each category at 50 images. This keeps any single
activity from dominating the benchmark and leaves 660 images.

\paragraph{Stage 4: human verification.}
Two annotators independently judge whether each image meets the singleton
constraint, and we keep only the images they agree on, leaving 538. These images
proceed to option generation and annotation
(Section~\ref{subsec:construction}); of them, 73 are spilt for the
fine-tuning analysis and the remaining 465 enter benchmark assembly. During
annotation, an image is dropped if none of its generated correct-answer
candidates survive verification (i.e., the annotators retain no appropriate
action for it), so the final benchmark draws on the 440 images that keep at least
one valid item.
\section{Annotation Agreement}
\label{app:iaa}

Each candidate is labeled independently by two annotators into one of three
categories: pass (a valid correct answer), off-vantage (a valid distractor), or
remove, and only candidates the two label identically enter the benchmark
(Section~\ref{subsec:construction}). Because off-vantage and remove lead to different
outcomes, namely a retained distractor versus a discard, we report the three-way
Cohen's $\kappa$ over these labels, which counts every cross-category
disagreement. Agreement is substantial, with $\kappa=0.61$ on COCO and
$\kappa=0.72$ on Wikimedia.

\paragraph{Per-label counts.}
Of the 2{,}291 candidates, 1{,}168 are kept as correct answers (767 direct, 401
indirect) and 257 as off-vantage distractors; the remaining 866 are discarded.
Of the discards, 523 ($23\%$ disagreement rate over all 2,291 candidates) are candidates on which the two annotators disagree
and 343 are candidates both agree to remove. Each retained correct answer becomes
one item, so the kept candidates yield 1{,}168 items before the image-level
hold-out. The off-vantage candidates provide the fourth option
for n4 items. The 257 off-vantage distractors consist of 230 candidates generated
as off-vantage and 27 reused from correct-answer candidates that both annotators
judged incompatible with the viewpoint (13 direct, 14 indirect).

\section{Out-of-Domain Set}
\label{app:wiki}
To examine the generalization of zero-shot and EAS tuning to diverse real-world
settings, we collected an out-of-domain set from Wikimedia through the same
construction pipeline (Section~\ref{subsec:construction}).
The pattern is consistent with the main results, and the models find these items
slightly more difficult than the COCO-based items. In zero-shot, open-source
models struggle substantially with this task: gemma4 is again the strongest at
$44.9\%$, while the others are below chance. The direct--indirect split persists,
with each model scoring much lower on direct items than on indirect items
(Table~\ref{tab:wiki}). The accuracy-improvement trend under EAS tuning is also
similar.

All Wikimedia Commons (\url{https://commons.wikimedia.org}) images used in this set are under permissive licenses (CC BY, CC0, or Public Domain). We release the Commons file-page URLs together with our annotations.
\begin{table}[t]
\centering
\caption{Out-of-domain (Wikimedia) accuracy (\%), sorted by overall accuracy
within each setting. The zero-shot direct--indirect split and the partial
recovery under rationale tuning reflect the main results
(Tables~\ref{tab:zeroshot} and~\ref{tab:rationale}). This set is small and is
reported as a qualitative check.}
\label{tab:wiki}
\small
\setlength{\tabcolsep}{8pt}
\begin{tabular}{l ccc cc}
\toprule
Model & Overall & n3 & n4 & Direct & Indirect \\
\midrule
\multicolumn{6}{l}{\emph{Zero-shot}} \\
gemma4     & 44.9 & 52.5 & 35.4 & 25.0 & 79.5 \\
internvl35 & 18.7 & 16.9 & 20.8 & 10.3 & 33.3 \\
qwen35     & 17.8 & 22.0 & 12.5 & \phantom{0}8.8 & 33.3 \\
qwen25vl   & 15.9 & 16.9 & 14.6 & 11.8 & 23.1 \\
internvl25 & 14.0 & 11.9 & 16.7 & 10.3 & 20.5 \\
\midrule
\multicolumn{6}{l}{\emph{EAS tuning}} \\
qwen35     & 65.4 & 78.0 & 50.0 & 50.0 & 92.3 \\
gemma4     & 48.6 & 57.6 & 37.5 & 30.9 & 79.5 \\
qwen25vl   & 44.9 & 49.2 & 39.6 & 38.2 & 56.4 \\
internvl25 & 36.4 & 37.3 & 35.4 & 30.9 & 46.2 \\
internvl35 & 34.6 & 32.2 & 37.5 & 22.1 & 56.4 \\
\bottomrule
\end{tabular}
\end{table}
\section{Training Details}
\label{app:training}

Both fine-tuning analyses use an identical training recipe and differ only in the
data. \emph{Egocentric tuning on EgoIT-99K} fine-tunes on standard egocentric
instruction data, and \emph{EAS tuning} fine-tunes on the EgoGapBench
split set; nothing else changes between them.

\paragraph{Data.}
For Egocentric tuning on EgoIT-99K we sample videos and images from EgoIT-99K~\citep{yang2025egolife},
drawing from seven of its constituent sources: Charades-Ego~\citep{sigurdsson2018charades}, EGTEA Gaze+~\citep{li2018eye}, EgoProceL~\citep{bansal2022my},
EPIC-KITCHENS, HoloAssist~\citep{wang2023holoassist}, IndustReal~\citep{schoonbeek2024industreal}, and Ego4D. We cap the sample at 800 examples per source, giving 5{,}272 examples in total for one epoch. These consist of 4{,}596 video-based examples and 676 image-based examples (Table~\ref{tab:trainmix}); Each example is a multi-turn dialogue, so the 5{,}272 examples
amount to roughly 16{,}900 question--answer turns (3.2 turns per example on
average). For EAS tuning we use a split of
EgoGapBench. The split is made at the image level, so that items generated from
the same image do not appear in both training and test, ruling out leakage
through shared scenes.

\paragraph{Optimization.}
We fine-tune with ms-swift (v4.2.3) using LoRA applied across the vision encoder
and the language model (\texttt{target\_modules}: all-linear; no modules frozen).
We keep the framework's default LoRA learning rate of $1\mathrm{e}{-4}$, with
rank $16$, $\alpha=32$, and dropout $0.05$. Training runs for one epoch with a
cosine schedule, warmup ratio $0.1$, an effective batch size of $16$ (per-device
batch $2$, gradient accumulation $8$), maximum sequence length $8192$, and bf16
precision. All runs use seed $42$.

\begin{table}[t]
\centering
\caption{Vision-tuning data mixture, the actual build from EgoIT-99K.
Six sources reach the per-source cap of 800; Ego4D falls short at 472. Videos
and images are trained jointly.}
\label{tab:trainmix}
\small
\setlength{\tabcolsep}{10pt}
\begin{tabular}{l ccc}
\toprule
Source & Video & Image & Total \\
\midrule
CharadesEgo  & 800 & \phantom{0}\phantom{0}0 & 800 \\
EGTEA        & 692 & 108 & 800 \\
EgoProceL    & 734 & \phantom{0}66 & 800 \\
EPIC-KITCHENS & 730 & \phantom{0}70 & 800 \\
HoloAssist   & 537 & 263 & 800 \\
IndustReal   & 631 & 169 & 800 \\
Ego4D        & 472 & \phantom{0}\phantom{0}0 & 472 \\
\midrule
\textbf{Total} & 4{,}596 & 676 & 5{,}272 \\
\bottomrule
\end{tabular}
\end{table}
\section{Effect of Explicit Exclusion Instructions}
\label{app:observeself}

A natural question is whether the failures observed on EgoGapBench can be
resolved simply by prompting. To test this, we append a
single instruction to the default prompt that states the rule explicitly, without
changing the options or the image:
\emph{``Do not choose an action that another person in the scene is performing;
instead, observe the scene and choose the action that is appropriate for you to
take.''}

Stating the rule does help. Accuracy rises substantially across all five
open-source models (Table~\ref{tab:observeself}), with gains on the direct items
that isolate EAS as well. However, even with this explicit hint, a large gap to
human-level performance remains, and for some models the effect is very limited.

\begin{table}[t]
\centering
\caption{Effect of the explicit instruction, which tells the model
not to choose another person's action, on the five open-source models. Subscripts
give the gain over the zero-shot prompt (Table~\ref{tab:zeroshot}).
Accuracy rises across all models but stays well below the human level (direct
$93.2\%$), and the gain is small for some models (e.g., InternVL2.5). Accuracy
(\%).}
\label{tab:observeself}
\small
\setlength{\tabcolsep}{6pt}
\begin{tabular}{l ccc cc}
\toprule
Model & Overall & n3 & n4 & Direct & Indirect \\
\midrule
gemma4     & 70.6$_{+18.9}$ & 77.2$_{+23.8}$ & 60.0$_{+11.0}$ & 56.2$_{+22.2}$ & 97.4$_{+12.8}$ \\
qwen35     & 57.1$_{+28.1}$ & 62.3$_{+30.4}$ & 48.7$_{+24.4}$ & 43.7$_{+23.7}$ & 82.0$_{+36.3}$ \\
qwen25vl   & 45.4$_{+16.8}$ & 48.4$_{+18.8}$ & 40.6$_{+13.6}$ & 35.7$_{+13.9}$ & 63.4$_{+22.3}$ \\
internvl35 & 41.9$_{+17.8}$ & 44.0$_{+18.9}$ & 38.5$_{+16.0}$ & 34.6$_{+14.1}$ & 55.4$_{+24.5}$ \\
internvl25 & 24.7$_{+8.6}$ & 24.3$_{+9.3}$ & 25.4$_{+7.6}$ & 24.0$_{+6.8}$ & 26.0$_{+12.0}$ \\
\bottomrule
\end{tabular}
\end{table}
\section{Prompt variation}
\label{app:egoplan}

To check that the performance gap is not an artifact of how the acting agent is
addressed or of the prompt style, we rerun the zero-shot evaluation under the
Egoplan-style prompt (Appendix~\ref{app:prompt}). The pattern of results is broadly
similar: gemma4 remains the clear top performer, every other open-source model
stays near or below chance, and the direct--indirect split is maintained
(Table~\ref{tab:egoplan}). The ranking is also moderately preserved (Spearman
$\rho=0.60$ against the default prompt): three of the five models keep their rank,
and only the two models clustered below chance swap places.

\begin{table}[t]
\centering
\caption{Zero-shot accuracy under the Egoplan-style prompt (\%). The below chance
performance and the direct--indirect split mirror the default prompt
(Table~\ref{tab:zeroshot}).}
\label{tab:egoplan}
\small
\setlength{\tabcolsep}{8pt}
\begin{tabular}{l ccc cc}
\toprule
Model & Overall & n3 & n4 & Direct & Indirect \\
\midrule
gemma4     & 49.4 & 53.2 & 43.2 & 32.9 & 80.0 \\
internvl35 & 28.2 & 29.3 & 26.4 & 17.7 & 47.7 \\
qwen25vl   & 27.9 & 28.3 & 27.2 & 22.5 & 38.0 \\
qwen35     & 21.8 & 24.6 & 17.3 & 14.0 & 36.3 \\
internvl25 & 19.4 & 18.9 & 20.2 & 16.8 & 24.3 \\
\bottomrule
\end{tabular}
\end{table}

\end{document}